\documentclass[conference]{IEEEtran}
\usepackage{amsmath,amsfonts}
\usepackage{amsthm}
\usepackage[linesnumbered,ruled,vlined]{algorithm2e}
\usepackage{array}
\usepackage[caption=false,font=normalsize,labelfont=sf,textfont=sf]{subfig}
\usepackage{textcomp}
\usepackage{stfloats}
\usepackage{url}
\usepackage{verbatim}
\usepackage{graphicx}
\usepackage{mathtools, nccmath}
\usepackage{cite}
\usepackage{balance}
\usepackage{booktabs}
\begin{document}
\title{Unsupervised Deep Unfolded PGD for Transmit Power Allocation in Wireless Systems}
\author{
Ramoni Adeogun\\
\textit{AI for Communications Group, Wireless Communication Network Section} \\
\textit{Department of Electronic Systems, Aalborg University, Denmark} \\
E-mail: ra@es.aau.dk
}
\maketitle

\begin{abstract}
Transmit power control (TPC) is a key mechanism for managing interference, energy utilization, and connectivity in wireless systems. In this paper, we propose a simple low-complexity TPC algorithm based on the deep unfolding of the iterative projected gradient descent (PGD) algorithm into layers of a deep neural network and learning the step-size parameters. An unsupervised learning method with either online learning or offline pretraining is applied for optimizing the weights of the DNN.  Performance evaluation in dense device-to-device (D2D) communication scenarios showed that the proposed method can achieve better performance than the iterative algorithm with more than a factor of $2$ lower number of iterations.
\end{abstract}


\IEEEpeerreviewmaketitle

\section{Introduction}

Transmit power control (TPC) has been studied extensively as a viable radio resource management technique for optimizing the performance as well as energy consumption of wireless systems, particularly in interference-limited scenarios. TPC typically involves the maximization (or minimization) of a communication theoretic objective function which is typically non-convex and NP-hard making deriving an optimal solution to become more difficult as the network size grows. Moreover, solutions must be obtained in real-time to adapt to the varying wireless channels. 

Due to the non-convexity and NP-hardness of power allocation, existing works have resulted in approximate solutions which rely on convex or geometric programming approximation of the problem. This has led to various approximate power allocation methods based on techniques such as particle swam optimization \cite{khodier2010beamforming}, game theory \cite{sengupta2009game}, and weighted minimum mean square error (WMMSE) algorithm \cite{SQWMMSE}. The applicability of these numerical or iterative solutions for real-time operation is, however, hindered by their high computation complexity, particularly for large-scale networks. Moreover, their performance is known to be sensitive to initialization and convergence may sometimes be too slow for real-time processing. 

Consequently, academic and industrial research focus appears to have shifted to machine learning-based algorithms with several published works applying graph neural networks \cite{shen2019graph, abode2022power, GNNDPower}, deep neural networks (DNNs) \cite{hameed2021deep}, reinforcement learning \cite{lu2018learning, zhang2020qlearning}, spatial DNN \cite{zhang2019energy} to the transmit power allocation problem in wireless networks. These learning-based solutions have been shown to achieve reasonably good performance with much lower complexity than classical iterative or heuristics algorithms for TPC. However, they are either based on the unrealistic assumption that network data for training machine learning models are available or that simulators can be developed to accurately model real wireless environments. In the wireless communication domain, the availability of suitable real data sets is still a challenge. Also, most of the existing machine learning-based methods suffer from slow convergence, non-interpretability, and/or overly complex network architecture.

To circumvent these limitations, there has been an increased interest in the deep unfolding of iterative algorithms \cite{unsupervised}, which aims to leverage the expressiveness and training capabilities of deep neural networks (DNNs) to improve performance with lower complexity compared to iterative algorithms. This approach allows for incorporating existing expert knowledge into the solution and has been applied to a wide range of optimization problems, including massive MIMO detection \cite{takabe2019trainable}, compressed sensing, sparse coding, and channel coding \cite{wadayama2019deep}. By unfolding the iterations of an iterative algorithm into a DNN, it becomes possible to optimize the algorithm through gradient descent, using backpropagation to adjust the DNN weights \cite{YoninaEldarDeepUnrolling}. This approach has been shown to improve convergence rates, reduce the required number of iterations, and allow for the incorporation of additional constraints or objectives into the solution. 

In this paper, we propose an unsupervised deep unfolded projected gradient descent (DUPGD) algorithm for power control in wireless networks. Our approach is based on deep unfolding the projected gradient descent algorithm, which is a widely used iterative algorithm for solving optimization problems. The DUPGD is trained following an unsupervised framework with the loss determined by the objective function in the TPC optimization problem. Performance evaluation results in device-to-device (D2D) communication scenarios have shown that DUPGD can achieve better performance than the classical iterative PGD for power allocation.  

The rest of this paper is organized as follows. The system model as well as problem formulation is presented in Section~\ref{sec:sysModel}. In Section~\ref{sec:proposedMethod}, the iterative PGD algorithm and the proposed DUPGD methods are described. Performance evaluation results are presented and discussed in Section~\ref{sec:perfEval}. Finally, conclusions are drawn in Section~\ref{sec:conc}.

\section{System Model and Proposed Algorithms}\label{sec:sysModel}
\subsection{System Model} 
We consider a typical device-to-device communication scenario with $N$ independent links (i.e., transmitter-receiver pairs) operating over a single shared frequency channel with bandwidth, $B$. The received signal-to-noise plus interference ratio (SINR) for the link between the $n$th transmitter and its receiver  $n'$ is defined as
\begin{equation}
    \zeta_{nn'} = \frac{p_nh_{nn'}}{\sum_{m=1,m\neq n}^N p_mh_{mn'}+\sigma^2},
\end{equation}
where $p_i$ denotes the transmit power of the transmitter, $i$, $h_{ij'}$ is the channel gain between transmitter $i$ and receiver $j'$ and $\sigma^2$ denotes the noise variance.

\subsection{Problem Formulation}
We consider the typical problem of maximizing the sum rate. The optimal transmit power vector $\boldsymbol{p} = [p_1,p_2,\cdots,p_N]^T$ can then be obtained as the solution of 
\begin{align}\label{eqOptimal1}
    \text{P}: &\quad \hat{\boldsymbol{p}} = \arg\hspace{-10pt}\min_{\boldsymbol{p}\in [0,p_{\mathrm{max}}]} -\sum_{n=1}^N \log_2\left(1+ \zeta_{nn'}\right)\nonumber\\
    & \quad\text{st:  }  0\leq p_n \leq p_{\rm{max}} \quad \forall n = 1,\cdots, N
\end{align}
where $p_{\rm{max}}$ denotes the maximum transmit power per transmitter. The $n$th term in the summation corresponds to the rate achieved by the $n$th transceiver pair.  We will henceforth use $\rho$ to denote the objective function in \eqref{eqOptimal1}. 
\section{Proposed Method}\label{sec:proposedMethod}
\subsection{Iterative PGD for Power Allocation}
To solve \eqref{eqOptimal1}, the PGD algorithm starts with an initial guess of the transmit power vector, $\boldsymbol{p}_0$, in the feasible set, i.e., $p_0\in [0,p_{\mathrm{max}}]$ and computes the iterate
\begin{equation}\label{PGD}
    \hat{\boldsymbol{p}}_{k+1} = \hat{\boldsymbol{p}}_{k} - \lambda \nabla \rho, \quad k = 1,\cdots, K 
\end{equation}
for a specified value of $K$ or until an appropriate termination criterion is met. In \eqref{PGD}, $\lambda (0\leq \lambda < 1)$ is the learning rate, and $\nabla \rho$ denotes the gradient of the negative sum rate $\rho$ with respect to the transmit power vector, $\boldsymbol{p}$, i.e.,  
\begin{align}
    \label{derivP2}
    \nabla \rho = \frac{\delta \rho}{\delta \boldsymbol{p}} = \left[\frac{\delta\rho}{\delta p_1}\quad \frac{\delta\rho}{\delta p_2}\quad \cdots\quad \frac{\delta\rho}{\delta p_N}\right]^T.
\end{align}
The derivative $\frac{\delta\rho}{\delta p_n}$ can be obtained using standard differential calculus rules to be 
\begin{equation}
    \label{derivP1}
    \frac{\delta\rho}{\delta p_n}= \frac{1}{N\log(2)}\left(\frac{h_{nn}}{\gamma_n}-\sum_{k=1,k\neq n}^N\frac{h_{kk}h_{kn}p_k}{(\eta_k)(\eta_k+h_{kk}p_k)}\right),
\end{equation}
where $\gamma_i = \sum_{m=1}^N p_mh_{im}+\sigma^2$ and $\eta_i = \sum_{m=1,m\neq i}^N p_mh_{im}+\sigma^2$. In PGD, the iterate in \eqref{PGD} is followed by projection onto the non-negative orthant and onto $p_{\mathrm{max}}$. A summary of the iterative PGD algorithm for TPC is shown in Algorithm~\ref{algo1}. Note that the convergence and the closeness of the power allocation in \eqref{PGD} to optimal solutions are highly dependent on the choice of the step-size parameter, $\lambda$. 

\begin{algorithm}[t]
\SetAlgoLined
\KwIn{Transmit powers $\boldsymbol{p}$, channel gains $\boldsymbol{h}$, noise variance $\sigma^2$, learning rate $\alpha$, number of iterations, $K$}
\KwResult{Optimal transmit powers $p^*$}
Initialize $p$ with some initial value\;
\For{$k = 1, 2, \ldots, K$}{
    Compute the $\nabla \rho$ with respect to $p$ using \eqref{derivP2}\;
    Update the transmit powers using \eqref{PGD}\;
    Project the transmit powers onto the feasible set: $\boldsymbol{p} = \text{proj}_{[0, P_{max}]^N}(\boldsymbol{p})$\;
}
\Return $\boldsymbol{p}^* = \boldsymbol{p}$\;
\caption{Iterative PGD for TPC.}\label{algo1}
\end{algorithm}
\begin{figure}[!t]
    \centering
    \includegraphics[width=1\columnwidth]{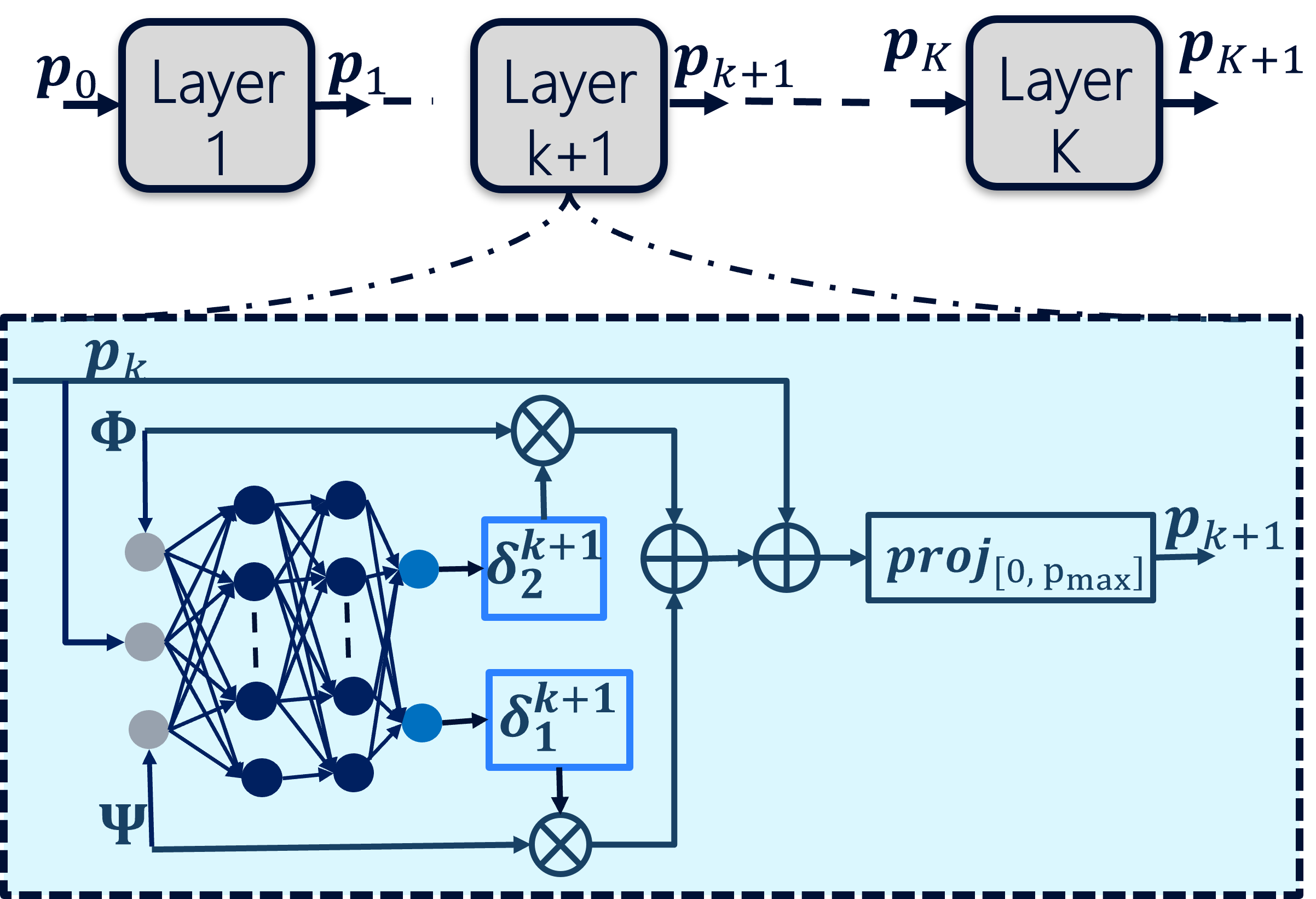}
    \caption{Illustration of proposed deep unfolded PGD for wireless power control.}
    \label{fig:DUPGD}\vspace{-10pt}
\end{figure}
\subsection{Deep Unfolded PGD for Power Allocation}
Deep unfolding is a technique that involves transforming an iterative optimization algorithm into a deep neural network (DNN) by unfolding the iterations of the algorithm into layers of the DNN. The resulting DNN approximates the optimal solution of the iterative algorithm.  We unfold the iterative PGD algorithm into a deep neural network (DNN) which can be trained offline translating to reduced execution complexity. By substituting \eqref{derivP1} and \eqref{derivP2} into the PGD iterate in \eqref{PGD}, we obtain
\begin{equation}\label{PGDE1}
    \hat{\boldsymbol{p}}_{k+1} = \hat{\boldsymbol{p}}_{k} - \lambda\boldsymbol{\Psi} + \lambda\boldsymbol{\Phi}, \quad k = 1,\cdots, K, 
\end{equation}
where 
\begin{equation}
 \boldsymbol{\Psi} = \frac{1}{\log(2)}\left[\frac{h_{11}}{\gamma_1}, \cdots, \frac{h_{NN}}{\gamma_N}\right]^T   
\end{equation}
 and 
 \begin{align}
   \boldsymbol{\Phi} &= \frac{1}{\log(2)}\left[\sum_{k=1,k\neq 1}^N\frac{h_{kk}h_{k1}p_k}{(\eta_k)(\eta_k+h_{kk}p_k)},  \cdots, \right.\nonumber\\ &\quad \quad\quad\quad\quad\quad\quad\left.\sum_{k=1,k\neq N}^N\frac{h_{kk}h_{kN}p_k}{(\eta_k)(\eta_k+h_{kk}p_k)}\right]^T.  
 \end{align}
 As seen in \eqref{PGDE1}, we require the terms $\boldsymbol{p}_k$, $\boldsymbol{\Phi}$ and $\boldsymbol{\Psi}$ to perform the $k+1$-th iteration of the PGD algorithm for TPC. These terms are then used as the input to the proposed deep unfolded PGD (DUPGD) model as shown in Fig.~\ref{fig:DUPGD}. Denoting the learned value of $-\lambda$ and $\lambda$ at iteration $k+1$ as $\delta_1^{k+1}$, $\delta_2^{k+1}$, the operations performed at the corresponding layer of DUPGD can be written as
 \begin{equation}
     \boldsymbol{p}_{k+1} = \operatorname{proj}_{[0,p_{\mathrm{max}}]}\left(\hat{\boldsymbol{p}}_{k} +\delta_1^{k+1}\boldsymbol{\Psi} + \delta_2^{k+1}\boldsymbol{\Phi}\right)
 \end{equation}

The overall structure of the DUPGD is depicted in Fig.~\ref{fig:DUPGD}. Unlike the iterative PGD method which relies on a manually specified value of the step-size parameter, $\lambda$, the step-size parameters are automatically determined in DUPGD via an unsupervised training procedure. In DUPGD, the step-size parameter is split into two learned parameters, $\delta_1^k \in [-1, 0]$ and $\delta_2^k \in [0,1]$ translating to an increased degree of freedom. At each layer of the DUPGD, the channel matrix, $\mathbf{h}$ and transmit power vector, $\boldsymbol{p}$ are converted into the vectors, $\boldsymbol{\Phi}$ and $\boldsymbol{\Psi}$ which are then passed as inputs to the layer as shown in Fig.~\ref{fig:DUPGD}. 

\subsubsection{Training procedure}
The DUPGD model is trained following the unsupervised training method described in \cite{unsupervised} with the loss function defined as the negative of the sum rate averaged over a batch of $N_B$ instantaneous channel gain matrices.  During training, the model is applied to predict transmit power levels for all nodes. The predicted transmit power is then used to update the weights of the DNN via backpropagation. In this, we used the Adam optimizer. It should, however, be noted that other stochastic gradient descent algorithms can be applied. We investigate two kinds of models viz: DUPGD with instantaneous online training (denoted DUPGD-online) and DUPGD with offline training (denoted DUPGD-offline), depending on how the training is performed. In DUPGD-online, the update of the model weights is performed layer-wise using a single instantaneous channel realization. In contrast, DUPGD-offline is trained prior to online execution with stochastic gradient descent over multiple batches of channel realizations. Once trained, the model is then deployed for online transmit power allocation translating to a lower execution complexity relative to DUPGD-online.
\begin{table}[!t]
\centering
\caption{Default simulation parameters}
\resizebox{0.95\columnwidth}{!}{
\begin{tabular}{@{}llll@{}}
\toprule
\textbf{Parameter}     & \textbf{Value} & \textbf{Parameter}           & \textbf{Value} \\ \midrule
Deployment area        & 20 m x 20 m    & Number of links, N           & 20             \\
Pathloss exponent      & 2              & Shadowing standard deviation & 5 dB           \\
Step size, $\lambda$       & 0.1            & Max number of iterations     & 1000           \\
Number of DUPGD layers & 40             & Number of neurons/layer      & 16             \\ \bottomrule
\end{tabular}}\label{tab:tab1}
\end{table}
\section{Performance Evaluation}\label{sec:perfEval}
\subsection{Simulation settings} 
We consider a D2D scenario with $N = 20$ links where the transmitters and receivers are randomly placed in a $20~\mathrm{m}\times 20~\mathrm{m}$ rectangular area. We consider deployments with no transmitter-to-receiver distance restriction and with a maximum transmitter-to-receiver distance of $d_{\mathrm{max}} = 3~\mathrm{m}$ which are henceforth denoted Scen 1 and Scen 2, respectively. While Scen 1 corresponds to a typical D2D communication scenario, Scen 2 represents scenarios where a transmitter communicates with a receiver in close proximity as in, e.g., short-range low-power 6G in-X subnetworks \cite{adeogun2022enhanced, adeogun2020towards} for extreme wireless communication inside entities such as the human body, industrial robots, and vehicles. 
Transmission bandwidth of $B = 5~\mathrm{MHz}$ with carrier frequency, $f_c = 6\mathrm{GHz}$ is considered in the simulations. With a noise figure of $10~\mathrm{dB}$, this translate to a noise power of $\sigma^2 = 2\times 10^{-10}$. Other simulation parameters are presented in Table~\ref{tab:tab1}. Except where stated otherwise, the maximum number of iterations for the iterative PGD method is set to $1000$. 
In the simulations, each layer of the DUPGD consists of a DNN with two hidden layers with 64 neurons per layer. Each hidden layer is followed by a rectified linear unit (ReLu) activation function. At the output layer, a clamping operation is performed to project the DNN prediction onto the feasible range between $0$ and $p_{\mathrm{max}}$.

 \begin{figure}
    \centering
\includegraphics[width=0.70\columnwidth]{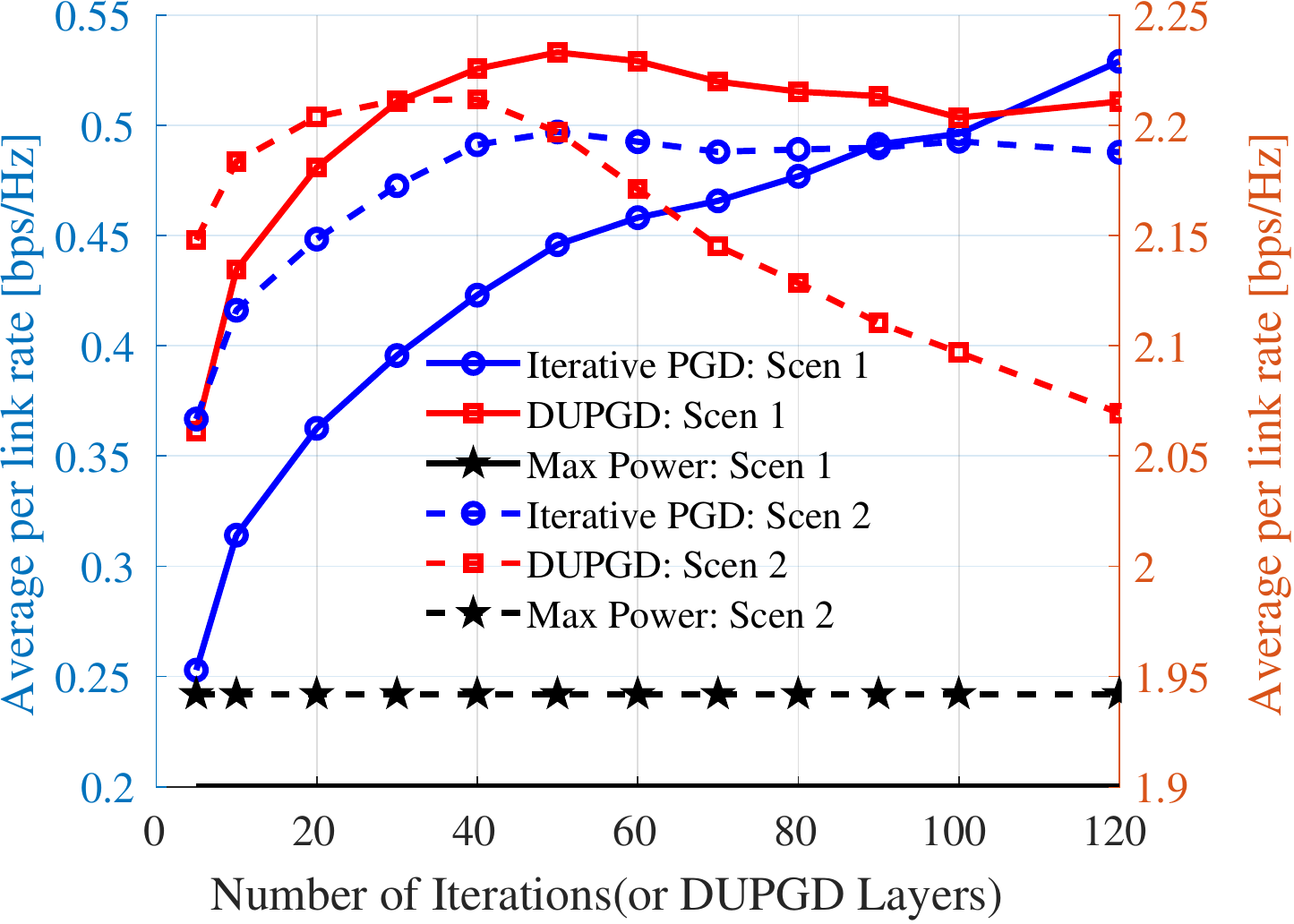}
    \caption{Average per link rate versus number of iterations with $N = 10$.}
\label{fig:rateversusIteration}\vspace{-5pt}
\end{figure}
\begin{figure}
    \centering
\includegraphics[width=0.82\columnwidth]{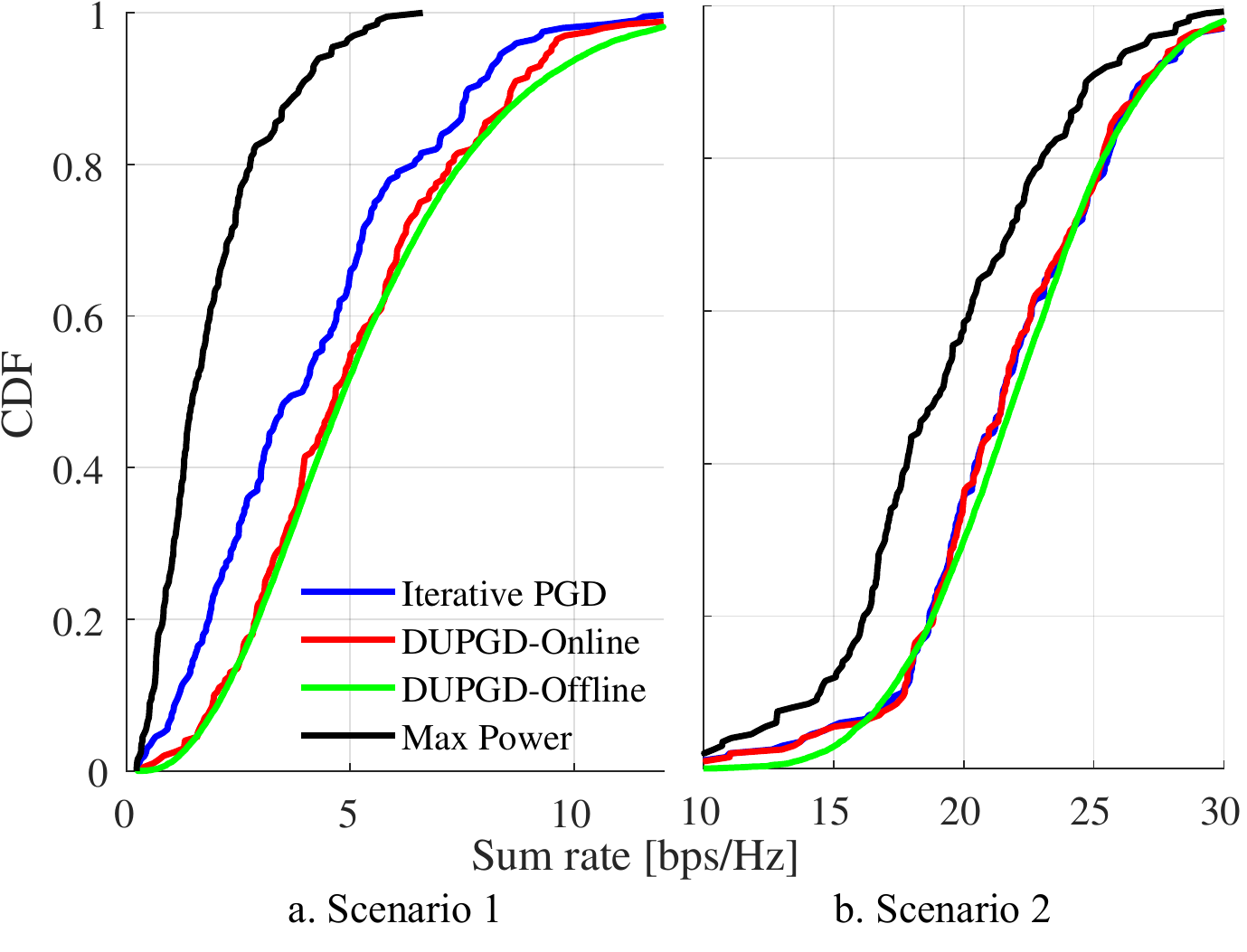}
    \caption{CDF of sum rate with number of links, $N = 10$.}
    \label{fig:sumRateCDF}\vspace{-5pt}
\end{figure}
\begin{figure}
    \centering    \includegraphics[width=0.7\columnwidth]{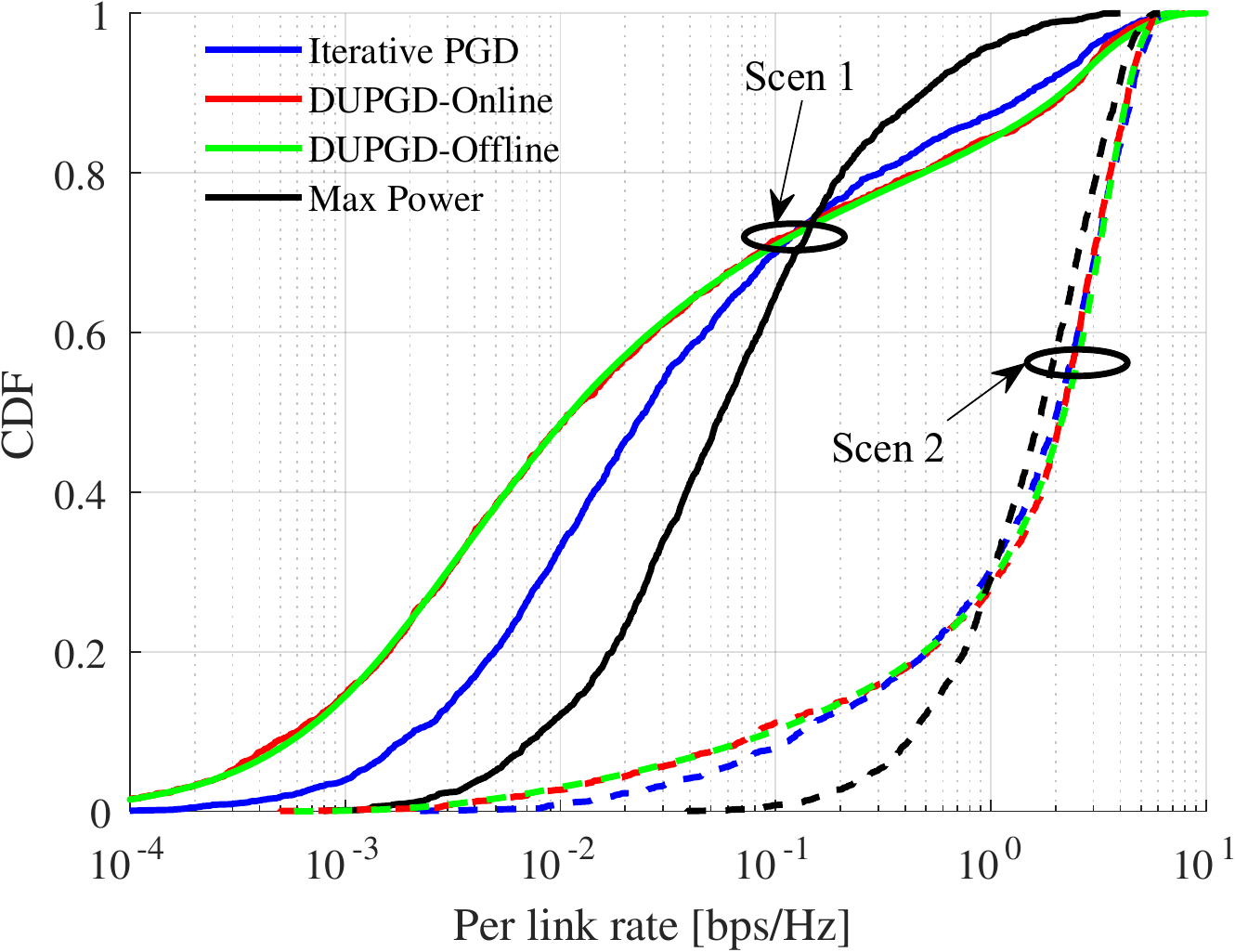}
    \caption{CDF of per link rate with number of links, $N = 10$.}
\label{fig:perlinkRateCDF}\vspace{-5pt}
\end{figure}

\begin{figure}
    \centering \includegraphics[width=0.7\columnwidth]{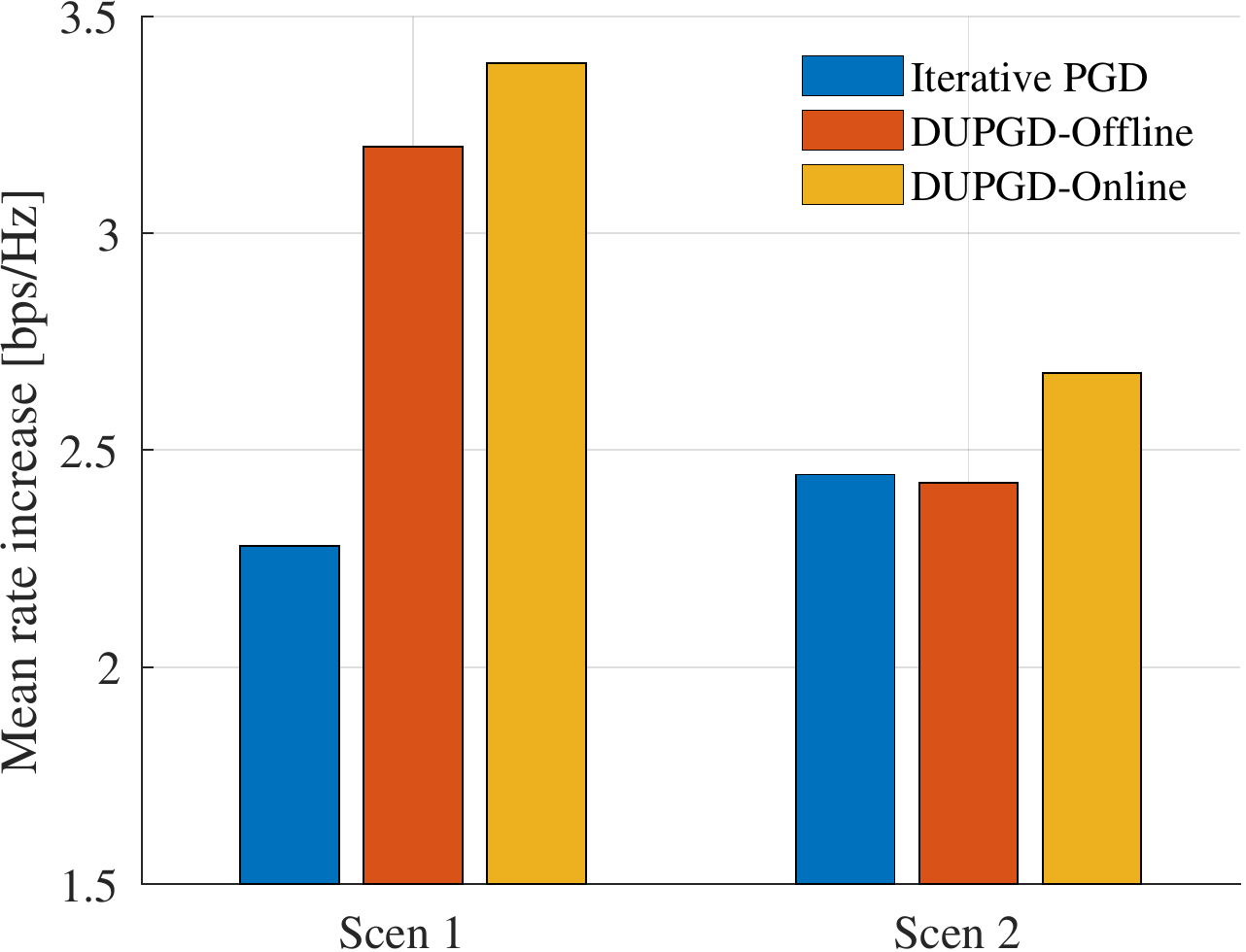}
    \caption{Mean rate improvement relative to the maximum power baseline with $N = 10$ links.}
\label{fig:meanIncrease}\vspace{-5pt}
\end{figure}
 \begin{figure}[!t]
    \centering
\includegraphics[width=0.75\columnwidth]{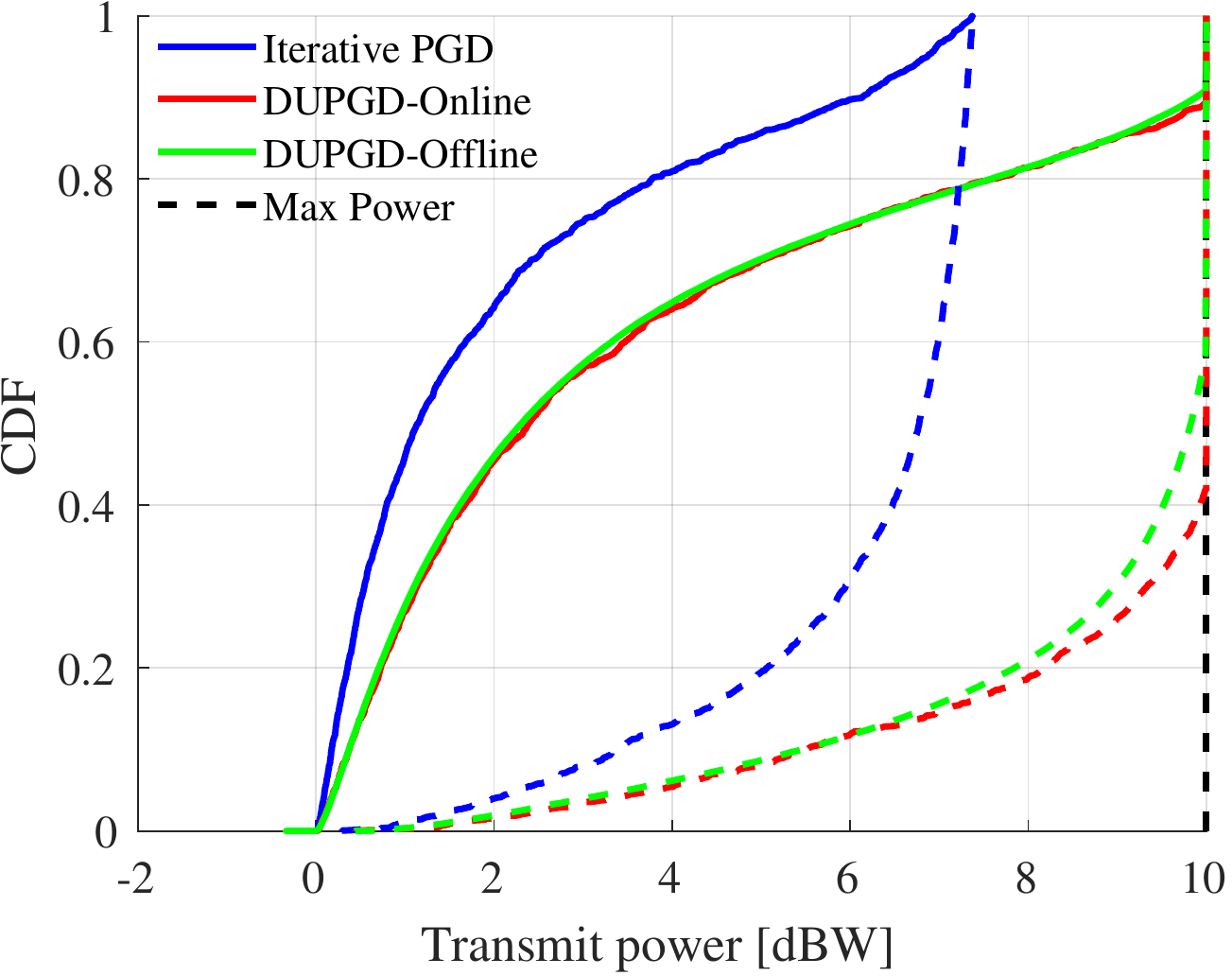}
    \caption{Distribution of allocated transmit power.}
    \label{fig:txPowerCDF}\vspace{-5pt}
\end{figure}

\subsection{Channel model}
The channel gain, $h$ is computed using 
\begin{equation}
    h_{ij'} = g_{ij'}\left(\frac{c^2\min (1, d_{ij'}^{-\omega})}{16\pi^2f_c^2}\Xi_{ij'}\right)^{\frac{1}{2}},
\end{equation}
where $g_{ij'}$ is small-scale fading gain generated as Rayleigh distributed random variables, $d_{ij'}$ is the distance between transmitter $i$ and receiver $j'$, and $\omega$ denotes the path-loss exponent. $f_c$, $c$, and $\Xi_{ij'}$ are respectively the carrier frequency, speed of light, and lognormal distributed shadowing of the link between the transmitter, $i$, and receiver $j'$. 

\subsection{Evaluation results} 
We evaluate the performance of the proposed DUPGD algorithm and compare it with that of classical iterative PGD and maximum power. In Fig.~\ref{fig:rateversusIteration}, we show the average per link rate as a function of the number of iterations (or equivalently the number of layers in DUPGD). As shown in the figure the average per-link rate (or equivalently the sum rate) increases with the number of iterations for both the DUPGD and iterative PGD. For the DUPGD methods, the maximum rate is achieved with the number of layers of approximately $40$. This indicates that a maximum of $40$ DUPGD layers are sufficient to achieve the best performance in the scenarios considered. In contrast, it appears that the iterative PGD's average rate performance remains constant after $60$ iterations for Scen 2 and continues to increase for Scen 1. It is therefore reasonable to conclude that the DUPGD requires a much less number of layers than the number of iterations in the iterative PGD. To minimize the effect of early termination on the performance achieved by the iterative PGD algorithm, we henceforth set the maximum number of iterations to $1000$.
Fig.~\ref{fig:sumRateCDF} shows the CDF of the sum rate achieved by the proposed and benchmark methods in scenario 1 (a) and scenario 2 (b). The figure shows that the DUPGD performs better than the equal power baseline in both scenarios and is similar in performance to the complex iterative PGD algorithm in scenario 2. In scenario 1, which is more challenging due to the completely random location of the transmitters and receivers, the DUPGD algorithm yields slightly better performance. The figure further shows no significant difference between the performance of DUPGD with offline and online training. This indicates that the proposed method can be used without any pre-training in cases where computation complexity associated with online training can be tolerated. In Fig.~\ref{fig:perlinkRateCDF}, we show the distribution of per link rate for the different algorithms. The results indicate that maximization of the sum rate as in iterative PGD and DUPGD leads to enhanced individual rates at the higher percentile of the CDF while penalizing the lower percentiles. However, this is expected considering that the optimization focuses only on the sum rate. In cases with a minimum rate requirement, the objective function can be augmented with a penalty term defined in terms of such a requirement. 

To quantify the relative performance improvement, we show the mean rate increase computed as the difference between the achieved rate by a specific algorithm and that achieved by the maximum power baseline in Fig.~\ref{fig:meanIncrease}. The figure shows that the proposed methods yield significantly higher mean rate increases of $3.39~\mathrm{bps/Hz} 
 (3.19~\mathrm{bps/Hz})$ for DUPGD-Online (Offline) compared to iterative PGD with an increase of $2.27~\mathrm{bps/Hz}$ in scenario 1. This shows that the proposed methods can achieve up to $\approx 49\%$ improvement in the relative rate increase compared to maximum power. The performance improvement in scenario 2 is only about $9.43\%$ (for DUPGD-Online) which is much lower than that achieved in scenario~1. A plausible explanation for this observation is that the relatively strong desired link power (due to the distance constraint) diminishes the impact of power control on the experienced aggregate interference level at each node. 
 
In Fig.~\ref{fig:txPowerCDF}, we show the distribution of the power allocated to each transmitter by the different methods. The figure shows that the iterative algorithm results in lower power consumption than the DUPGD in the considered scenarios. For instance, while the no transmitter gets the maximum power with the iterative algorithm, approximately $10\%$ and $60\%$ are allocated the maximum power of $10~\mathrm{dBW}$ with DUPGD in scenarios 1 and 2, respectively. The mean transmit power with iterative PGD and DUPGD are  $2.69~\mathrm{dBW}$ and  $5.21~\mathrm{dBW}$ (in scenario 1) and $6.31~\mathrm{dBW}$ and $9.27~\mathrm{dBW}$ (in scenario 2). 
\begin{figure}
    \centering    \includegraphics[width=0.71\columnwidth]{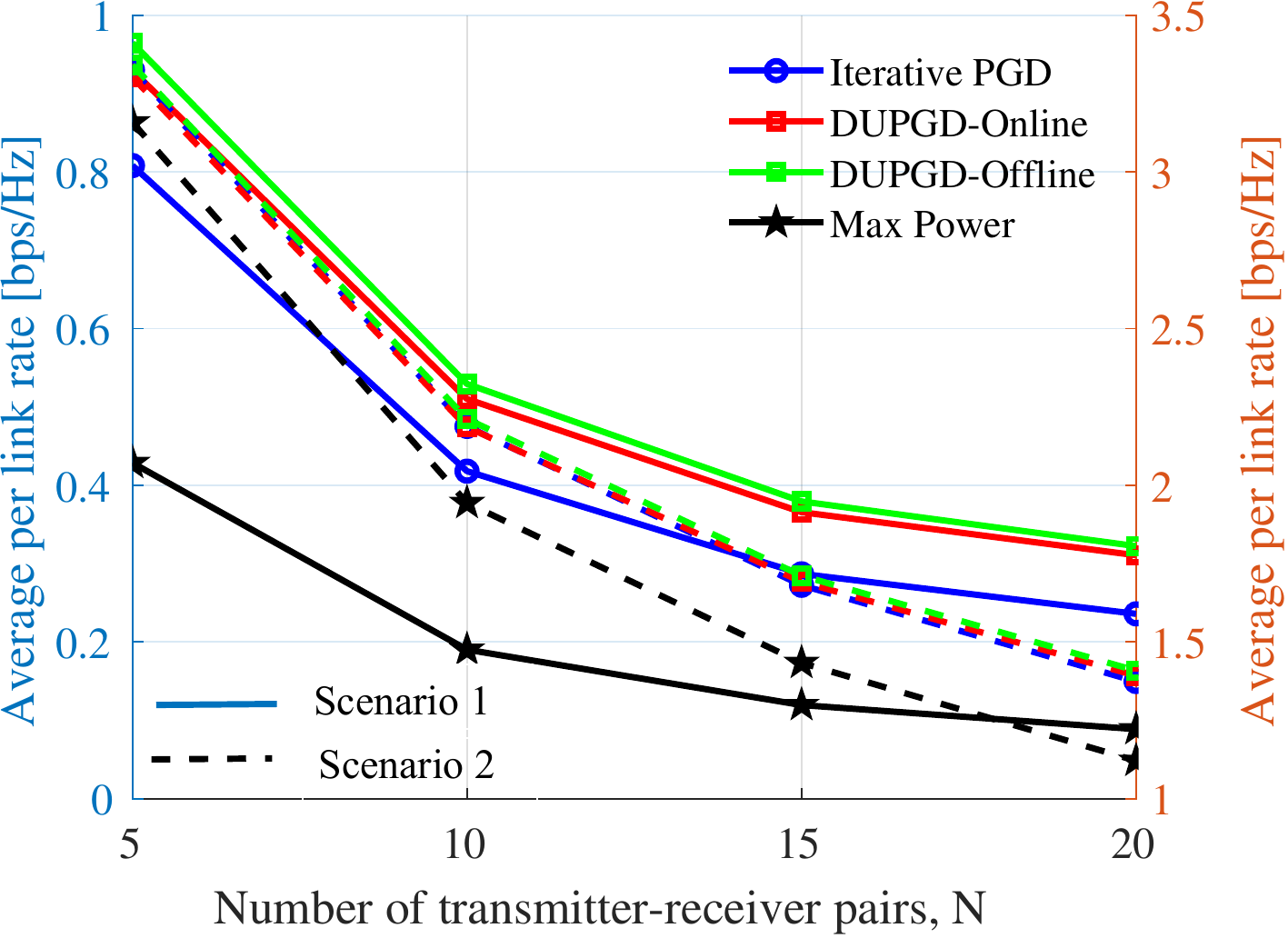}
    \caption{Averate rate versus the number of links.}
\label{fig:rateversusN}\vspace{-10pt}
\end{figure}
 \begin{figure}[!t]
    \centering    \includegraphics[width=0.71\columnwidth]{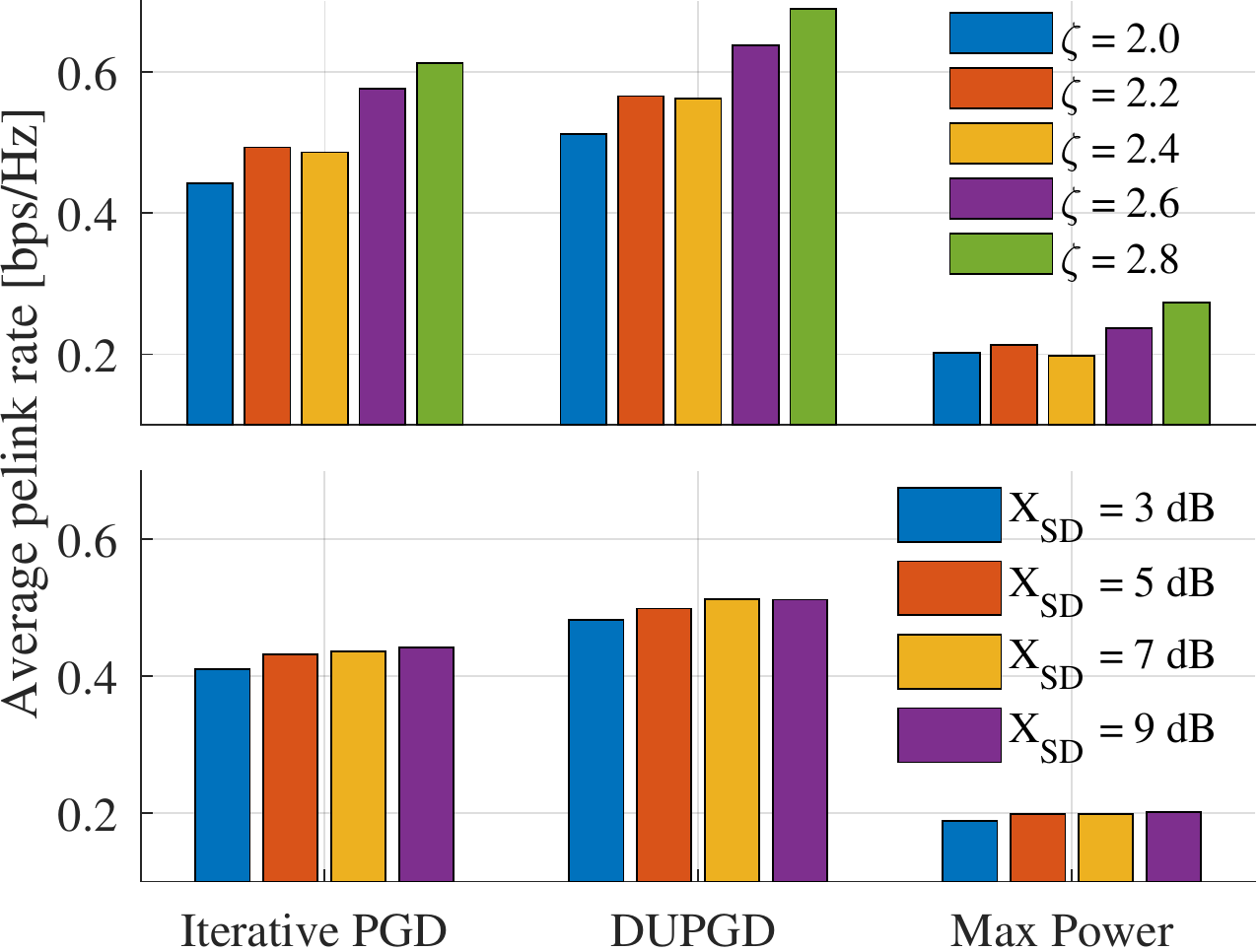}
    \caption{Sensitivity of TPC algorithms to changes in path loss and shadowing.}
\label{fig:sensPropagtion}\vspace{-10pt}
\end{figure}
 \subsection{Sensitivity evaluation} 
 We study the sensitivity of the proposed DUPGD method to changes in deployment density (or equivalently the number of subnetworks with fixed deployment area) and propagation conditions. In Fig.~\ref{fig:rateversusN}, we show the average per link rate as a function of the number of transmitter-receiver pairs, $N$, in a $20~\mathrm{m}\times 20~\mathrm{m}$ area. The figure shows that the relative performance of the proposed DUPGD to the baseline iterative algorithm remains constant for all values of $N$. This indicates that the DUPGD generalizes well to different deployment densities. Fig.~\ref{fig:sensPropagtion} shows the sensitivity of the power allocation methods to changes in the path-loss exponent and shadowing standard deviation. 
 The figure shows consistent relative performance between the DUPGD method and the benchmark algorithms for all values of path-loss exponent and shadowing standard deviation. It is therefore reasonable to conclude that the proposed method is robust to changes in wireless environment conditions.


\section{Conclusion}\label{sec:conc}
We proposed a low-complexity power control algorithm based on a deep unfolding of iterative project gradient descent. We showed via extensive simulations in device-to-device communication scenarios that the proposed method can achieve better performance than the classical iterative algorithm and decreases the number of iterations by more than a factor of $2$. Simulation results have shown that the proposed algorithm exhibits similar performance relative to the benchmark methods in environments with different propagation characteristics and deployment densities.  

\section*{Acknowledgement}
This research is partially supported by the HORIZONJU-
SNS-2022-STREAM-B-01-03 6G-SHINE project (Grant
Agreement No. 101095738).

\bibliographystyle{IEEEtran}
\bibliography{IEEEabrv,WPC}

\end{document}